\documentclass{article}

\usepackage[preprint]{neurips_2026}
\usepackage{amsmath}
\usepackage{caption}
\usepackage{amssymb}
\usepackage{wrapfig}   
\usepackage{float}     

\usepackage[utf8]{inputenc} 
\usepackage[T1]{fontenc}    
\usepackage{hyperref}       
\usepackage{url}            
\usepackage{booktabs}       
\usepackage{amsfonts}       
\usepackage{nicefrac}       
\usepackage{xspace}
\usepackage{microtype}      
\usepackage{xcolor}         
\usepackage{wrapfig}
\usepackage{booktabs}
\usepackage{multirow}
\usepackage{arydshln} 
 \usepackage{graphicx} 
\usepackage{booktabs}
\usepackage[table]{xcolor}
\usepackage{colortbl}
\usepackage{newunicodechar}
\usepackage{algorithm}
\usepackage{algpseudocode}
\usepackage{amsmath, amssymb}
\newunicodechar{↳}{\ensuremath{\hookrightarrow}}
\definecolor{groupgreen}{RGB}{220,245,225}
\definecolor{groupblue}{RGB}{225,235,255}
\definecolor{gtgreen}{RGB}{190,225,190} 
\definecolor{commentgreen}{RGB}{120, 200, 150}
\newcommand{\ours}{\textit{{ELM}}\xspace}
\newcommand{\myparagraph}[1]{\noindent\textbf{#1}}
\title{Continual Self-Improvement with \\  Lightweight Experiential Latent Memories }
\author{%
  Vaggelis Dorovatas$^{1}$ \\
  \And
  Nancy Kalaj$^{1, 2}$ \\
  \And
  Rahaf Aljundi$^{1}$ \\
  \AND
  $^{1}$\normalfont Toyota Motor Europe \\
  $^{2}$\normalfont University of Trento \\
}

\begin{document}

{
\renewcommand{\thefootnote}{}
\footnotetext{*First author provides contracted services for Toyota.
}
\footnotetext{*Correspondence: vdorovatas@hotmail.gr}
}
\maketitle

\begin{abstract}
Large language models achieve strong reasoning performance by scaling inference-time compute, yet remain fundamentally stateless, discarding the rich, self-produced reasoning traces generated during this process. We investigate whether models can instead learn online from this experience, converting transient computation (reasoning traces) into persistent reusable knowledge, and without external supervision or access to future data. We show that In-Context Learning (ICL) over raw reasoning traces fails to generalize, reflecting a fundamental limitation of token-level reuse: individual traces lack the abstraction needed for transfer, even after refinement (e.g. self-reflection). In contrast, drawing inspiration from recent works on unsupervised reinforcement learning, we find that lightweight per-instance training with self-generated test-time signals (majority voting) as rewards yields substantial gains, often surpassing full-dataset offline training, motivating a shift from raw traces to learned latent representations. Building on this insight, we propose an online method that distills inference-time compute spent on encountered problems into compact modular latent memories capturing the underlying reasoning structure. These memories are stored and retrieved for future inputs, enabling continual improvement while avoiding catastrophic forgetting through modular design. Importantly, our method is highly efficient, parametrized as extremely lightweight soft prompt memories ($\sim$
0.001\% of model parameters) and trained with only a few gradient steps, yet achieving performance competitive with full parametric updates and offline training. Across challenging mathematical reasoning benchmarks, our approach significantly outperforms zero-shot and raw data ICL baselines, while transfering effectively across datasets.
\end{abstract}

\section{Introduction}
Large language models (LLMs) have demonstrated remarkable reasoning capabilities (e.g., ~\cite{jaech2024openai,guo2025deepseek, shao2024deepseekmath}), increasingly through extended inference-time compute and thinking~\cite{wei2022chain,snell2025scaling,brown2024large}, yet they remain stateless after deployment, spending significant compute on each problem without ever accumulating experience or converting the compute spent into reusable knowledge for the future. As models are increasingly deployed in interactive, problem-solving settings, the reasoning traces they produce represent a rich and largely untapped source of self-generated data that the model can exploit to learn continuously. A natural question arises, how an LLM can self-improve online and learn from its experience by converting its own thinking process into reusable knowledge (we refer to hereafter as experiential memories). 
Recent work has pursued this question primarily through training-free methods based on in-context learning (ICL)~\cite{suzgun2025dynamic,anonymous2026agentic,ouyang2025reasoningbank}, which store and retrieve past reasoning traces to inform new predictions. While appealing in their simplicity, these works force reasoning traces storage through a textual bottleneck,  by storing only what a model \emph{said} rather than what it \emph{computed}, discarding rich information about intermediate states, explored dead ends, confidence, and structural understanding. Recent work on latent reasoning~\cite{hao2024training}
demonstrates that models can “think” in continuous latent space more efficiently than in discrete
token space,  probing studies~\cite{belinkov2017neural,geva2021transformer} show that crucial information exists in hidden states that never surface to the text output. Chain-of-thought faithfulness work~\cite{lanham2023measuring} reveals that CoT explanations often do not reflect actual model computations.
We further show that current training-free approaches for experiential memories yield negligible improvements on challenging mathematical benchmarks. This failure reflects a fundamental limitation of storing reasoning traces or their textual abstraction: retrieving a  past reasoning trace provides a sample of prior experience but not a meaningful learning signal, yielding shallow generalization. We thus posit, and empirically validate, that meaningful memories should instead be learned from the model's explored trajectories, successes and failures when trying to solve a given problem with test-time compute, thus internalizing the findings of the reasoning process rather than storing its surface-level output.
\begin{figure}
    \centering
    \includegraphics[width=1.0\linewidth]{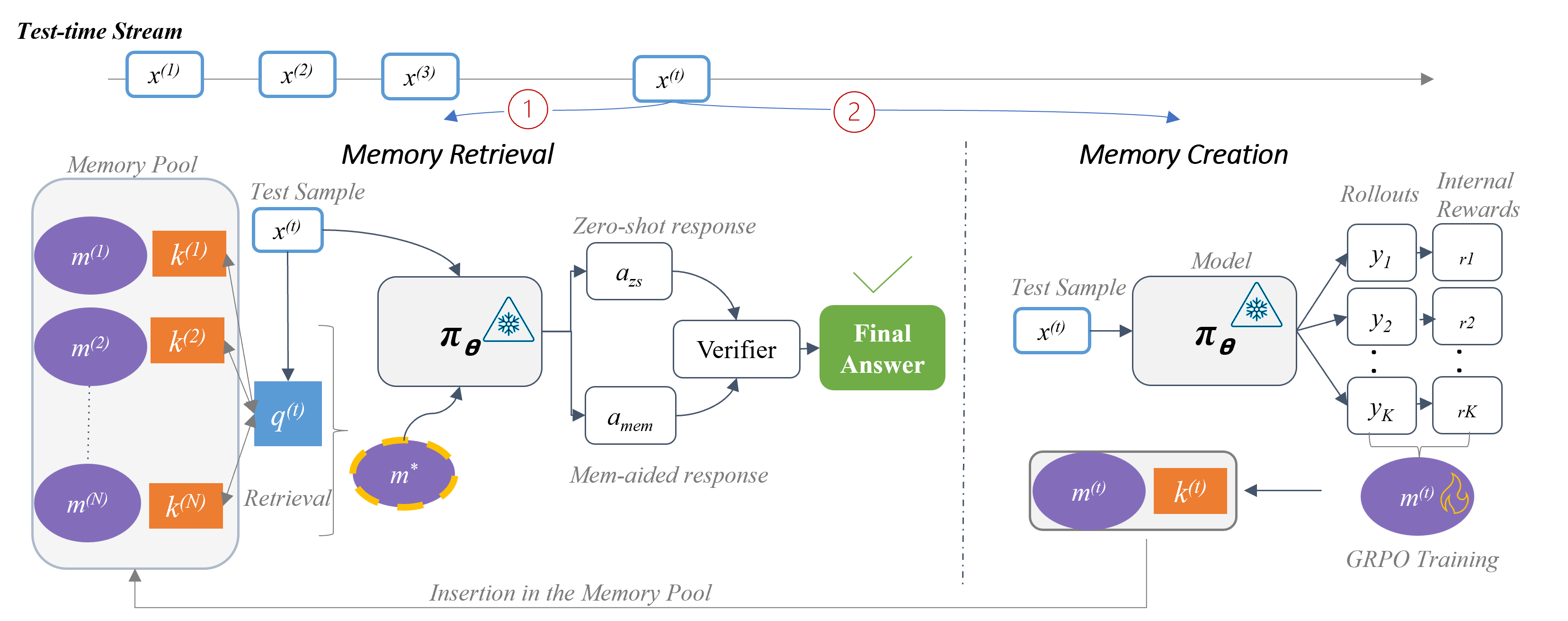}
    \caption{\small\textbf{Experiential Latent Memory} for continual self-improvement from a test-time stream.  Memory retrieval: given a test sample, we query the most relevant memory. For reliability, a verifier routes between zero-shot and memory-augmented responses, selecting the higher-scoring output. 
 Memory creation: whenever possible, we use additional test-time compute to train a lightweight modular memory for a given test sample with GPRO and internally derived rewards. Newly created memories are added to the memory pool for later reuse.}

    \label{fig:teaser}
\end{figure}

Another relevant line of work uses Reinforcement Learning (RL) with self-generated (internal) reward signals to drive improvement without ground-truth labels, demonstrating meaningful gains from purely internal supervision~\cite{zuo2025ttrl, zhao2025learning,jayalath2025compute, zhao2025absolute}. However, these approaches operate in an offline regime: they require access to the full test dataset upfront, training on it before gains on this set can be shown. Beyond this practical limitation 
and even more importantly, such methods have been recently shown to degrade as 
test-time data accumulate, essentially showing that they do not scale~\cite{zhang2025no,he2026far} possibly due to interfering self-supervisory signals. A final drawback is that these approaches rely on full-model updates, potentially leading to forgetting, optimization instability, and substantial compute and memory overhead. The truly online streaming setting, where the model encounters each problem once and must efficiently improve continuously as experience accumulates 
and without prior data access, remains largely unexplored and unsolved. To the best of our knowledge, this is the first work that addresses the question of \emph{ how can an LLM internalize its experiences and reasoning explorations without external supervision in a streaming fashion, enabling steady self-improvement with no interference or forgetting of its existing capabilities?}

We begin with a key empirical observation: training on individual examples for a few steps using RL with internal rewards can substantially boost model's performance (generalization), even outperforming full-dataset offline training, especially as we increase the amount of data used for the offline  training. In other words, when leveraging internal model feedback, learning from reasoning traces of individual  samples can elicit generalization that surpasses full offline training. 
Importantly, we show that introducing modular and efficient parameterization that leaves the main model intact can further improve the performance, while significantly reducing compute and memory requirements.  
To this end, we introduce an efficient online continual self-improvement method in which the model, after solving each problem, constructs a lightweight  latent memory, parametrized as a soft prompt, distilling the test-time compute spent on this experience into a compact latent representation stored and retrieved for relevant future inputs. 

Our approach addresses limitations of both lines of work. Unlike raw-data ICL, latent memories operate in the model’s representational space, capturing abstract reasoning rather than surface patterns, leading to stronger generalization. Unlike offline RL methods, \ours is designed for streaming settings and scales with accumulated experience, mitigating forgetting via modularity: memories are isolated and composed at retrieval time without interference. 
Our method is also highly efficient, using lightweight memories ($\sim$0.001\% of parameters) trained with few gradient steps, yet achieving performance competitive with full fine-tuning. Fig.~\ref{fig:teaser} illustrates \ours (Experiential Latent Memory). Empirically, \ours outperforms zero-shot and ICL baselines on challenging math benchmarks, shows strong cross-dataset transfer, and provides a scalable path to continual self-improvement by converting test-time compute into reusable knowledge.
Our contributions are as follows:  1) We introduce latent experiential memories for efficient, modular instance-level learning. 2) We integrate them into an online protocol with per-sample storage and retrieval. 3) We demonstrate strong gains on challenging benchmarks with minimal overhead.

\section{Related Work}
\myparagraph{RL for LLM Reasoning.}
Reinforcement learning with verifiable reward (RLVR) improves LLM reasoning via rule-based rewards~\cite{gao2024designing} or answer matching against ground truth~\cite{zeng2025simplerl}. While typically applied at scale, recent work shows that strong reasoning can emerge from few examples, supported by a theoretical lower bound on the number of required training instances~\cite{zhang2026resource} and empirical results with small or even single training instances~\cite{fatemi2025concise,wang2025reinforcement}. However, these works do not provide mechanisms for selecting or exploiting such examples. 
\\
To reduce reliance on ground-truth labels, prior work explores internally derived rewards~\cite{xu2025genius,zhao2025absolute,jayalath2025compute,zuo2025ttrl,zhao2025learning}. Closest to ours, TTRL~\cite{zuo2025ttrl} generates rewards via majority voting, however, test-time data are collected offline and used to train the full model with GRPO~\cite{shao2024deepseekmath}. In contrast, our approach demonstrates that efficient online training on test samples can yield performance gains that rival offline test-time training variants.\\
\myparagraph{Memory of Past Experience for Improved Reasoning.}
Prior work stores test-time experience, typically as text~\citep{suzgun2025dynamic,anonymous2026agentic,ouyang2025reasoningbank}, often requiring ground-truth supervision and large models. This textual bottleneck discards rich internal signals (e.g., intermediate states, uncertainty). LAG~\cite{chen2025log} instead reuses KV activations, but remains limited without learning. Learned latent memory methods (e.g., MemGen~\cite{zhang2025memgen}, FlashMem~\cite{hou2026flashmem}) rely on offline-trained modules and remain largely static at test-time. 
In contrast, we learn latent memories directly from individual experiences using internal feedback, enabling continual online adaptation and improved generalization.
\\
\myparagraph{Online Continual Learning.}
Our work is related to online continual learning~\citep{aljundi2019gradient}, where models learn from a stream of data while accumulating knowledge over time. Classical approaches update a single model under supervised streams, where the main challenge is catastrophic forgetting of previously learned knowledge~\citep{french1999catastrophic}. 
A complementary line considers self-supervised continual learning, where supervisory signals are internally generated~\citep{liang2025comprehensive}. Unlike these approaches, which still update a single model, we address forgetting by introducing experiential latent memories: experience is distilled into modular, individually parameterized units that are retrieved at test time. This design mitigates interference and enables compact, scalable knowledge accumulation, aligning with recent views on modular long-term memory~\citep{dorovatas2026modular}.

\section{\ours: Continual Self-Improvement with \emph{E}xperiential \emph{L}atent \emph{M}emories}
Our goal is to enable a model of continuous self-improvement by distilling test-time compute into reusable experiential memories. These memories can be accumulated and retrieved to improve performance on relevant future test samples with less compute. 
To approach this goal, one has to answer two questions: (1) how to learn \emph{efficiently and effectively} from a given sample using internal model feedback, and (2) how to leverage previously constructed memories to elevate performance on upcoming test samples.
In the following, we describe our approach and key design choices.
\subsection{Preliminaries} 

\paragraph{Group Relative Policy Optimization (GRPO).}

GRPO~\citep{shao2024deepseekmath} is a policy gradient method that eliminates the need for a separate critic by estimating advantages relative to a group of sampled outputs. Given a question $x$,  a group of $G$ responses $\{y_1, \dots, y_G\}$ are sampled from the current policy $\pi_\theta$, and the advantage of each response is computed by normalizing rewards within the group:
\begin{equation}\label{eq:grop}
    \hat{A}_i = \frac{r_i - \text{mean}\!\left(\{r_j\}_{j=1}^G\right)}{\text{std}\!\left(\{r_j\}_{j=1}^G\right)},
\end{equation}
where $r_i$ is the reward assigned to response $y_i$. The policy is then updated by maximizing a clipped surrogate objective with KL regularization to the initial
policy. 

\paragraph{Test-Time Reinforcement Learning (TTRL).}
TTRL~\citep{zuo2025ttrl} suggests an internal reward in  the setting where ground-truth labels are unavailable at test time, by leveraging additional test-time compute. 
Given a question $x$, the model generates $N$ candidate responses $\{y_1, \dots, y_N\}$, and a consensus answer $y^*$ is derived via majority voting, serving as a proxy label for reward computation without any external supervision:
\begin{equation}\label{eq:ttrl}
    R(\hat{y}_i, y^*) = \begin{cases} 1 & \text{if } \hat{y}_i = y^*, \\ 0 & \text{otherwise,} \end{cases}
\end{equation}
These self-produced binary rewards are then used for training with GRPO.

\subsection{How to Design Efficient and Effective Latent Memories?}

\begin{wrapfigure}{r}{0.5\textwidth}
\vspace*{-0.8cm}
\begin{minipage}{0.5\textwidth}
\scriptsize
\begin{algorithm}[H]
\caption{Continual Self-Improvement with Experiential Latent Memories (\ours)}
\label{alg:online_memory}
\begin{algorithmic}[1]
\State \textbf{Input:} Stream $\{x_i\}$, frozen policy $\pi_\theta$
\State \textbf{Initialize:} memory pool $\mathcal{M} \gets \emptyset$
\State \textbf{Hyperparameters:} \#rollouts $K$, train steps $S$
\For{each incoming sample $x_i$}
    \State \textit{\color{commentgreen} \# Memory Retrieval}
    \If{$|\mathcal{M}| > 0$}
        \State $q_i = \mathbf{Embed}(x_i)$
        \State $m^* \gets \operatorname*{\mathbf{Retrieve}}(\mathcal{M}, q_i)$
        \State \textit{\color{commentgreen} \# Response Selection}
        \State $a_{\mathrm{zs}} \sim \pi_\theta(\cdot \mid x_i)$
        \State $a_{\mathrm{mem}} \sim \pi_\theta(\cdot \mid [m^*, x_i])$
        \If{$a_{\mathrm{zs}} = a_{\mathrm{mem}}$}
            \State $\hat{y}_i \gets a_{\mathrm{zs}}$
        \Else
            \State $\hat{y}_i \gets \mathrm{Verifier}(a_{\mathrm{zs}}, a_{\mathrm{mem}})$
        \EndIf
    \Else
        \State $\hat{y}_i \sim \pi_\theta(\cdot \mid x_i)$
    \EndIf
    \State \Return $\hat{y}_i$ \quad 
    \textit{\color{commentgreen}\# Response Returned}
    \\
    \State \textit{\color{commentgreen} \# Memory Creation}
    \State  $m^{(i)} \gets \mathbf{Init}(x_i)$
    \State $m^{(i)} \gets \mathbf{GRPO}(m_{x_i}, K, S)$
    \State $\mathcal{M} \gets \mathcal{M} \cup \{m^{(i)}\}$
\EndFor
\end{algorithmic}
\end{algorithm}
\end{minipage}
\vspace*{-0.5cm}
\end{wrapfigure}
To build an effective online continual self-improvement method, we must ensure that memory representations and training signals can extract meaningful experience from encountered problems that is useful for future inquiries.
Recent work on the interplay between data and RL~\citep{wang2025reinforcement} has shown that individual samples exist within a training set such that, when trained on, they yield significant gains on the broader test set.
This motivates our assumption that training on individual samples encountered at test-time can provide strong benefits for future inquiries if employed correctly. However, 
for self-improvement, one has to extract internal training signal without  explicitly relying on external feedback, which is often unavailable or scarce in practice. 
Thus, given a test-sample $x$ and a current policy $\pi_\theta$ we adopt TTRL~\cite{zhao2025learning,zuo2025ttrl}, in which multiple samples are drawn to predict a consensus answer $y^*$ via majority voting. An internal reward is then  extracted as illustrated in Eq.~\ref{eq:ttrl} and GRPO is used for training (Eq.~\ref{eq:grop}).



However, leveraging this training objective to update the full model as test samples are received is not realistic for two main reasons:
(1) online updates to the full model for each sample incur substantial computational costs, rendering such an approach impractical; and
(2) more importantly, continuous updates on a stream of samples risk catastrophic forgetting~\citep{french1999catastrophic} of the model’s initial capabilities and previously acquired experience.
\\
In this work, to address these two challenges, we posit that experience acquired at test time need not be injected directly into the full model. Instead, we propose a modular design in which experience from each test sample is captured independently and stored for later reuse in the form of a modular \emph{experiential memory}.
By isolating the experience acquired from each sample in an independent memory, we ensure stability of the main model and, more importantly, minimize interference between memories derived from individual samples. Furthermore, to enable effective reuse, these memories must be retrievable—i.e., stored in a form that allows them to be efficiently recalled given future samples that can benefit from the distilled experience.
To this end, we employ soft prompts~\citep{lester2021power} as the parameterization of our experiential memory.

\paragraph{Soft Prompts as Latent Experiential Memories.}
Soft prompts~\citep{lester2021power} are sequences of trainable continuous vectors prepended to the input embeddings, which are optimized directly in the model’s latent space while keeping all other parameters frozen. Formally, given an input $x$, the effective input to the model becomes$[p_1, \dots, p_k, x]$, where $\{p_i\}_{i=1}^k \in \mathbb{R}^d$  are free parameters optimized via gradient descent and $d$ denotes the model’s embedding dimension. This defines a memory $m_x \in \mathbb{R}^{k \times d}$ that captures the experience acquired from solving sample $x$.
The choice of soft prompts for capturing experience as modular memory is motivated by two key properties. (1) Efficiency: the memory is parameterized as an extremely lightweight soft prompt (approximately ${\sim}0.001\%$ of the total model parameters), allowing the base model to remain frozen throughout. (2) Retrieval utility: the memory is initialized from the sample’s textual embedding and refined through a small number of update steps in the model’s latent space, enabling meaningful similarity comparisons with embeddings extracted from future queries.

\paragraph{Memory Creation.}
For each encountered input, the model first generates a solution using the current policy. We then distill this experience into a memory by lightly training for few gradient steps using GRPO~\ref{eq:grop}, with majority-voting rewards~\ref{eq:ttrl} serving as internal supervision. This process requires no ground-truth labels and keeps the main model frozen, thereby maintaining a stable policy reference and minimizing the capture of spurious noise in these self-supervised memories.
Each memory created from a given test sample represents a distilled unit of experience. 
\\
In the experiments, Sec.~\ref{sec:offline_experiments}, we analyze the generalization capability of individual memories in comparison to offline training of the full model, and show that: (1) individual memories can drive strong generalization that rivals offline full-dataset training; (2) the utility of memories varies across samples and depends on the nature of the experience captured; and (3) Noisy memories that degrade generalization performance exist, and handling them is essential for reliable continuous self-improvement.
\\
We hypothesize that, in full offline training, signals from both noisy and beneficial outcomes are mixed, which can lead to averaged performance gains as well as instability or collapse effects that have been reported as potential drawbacks of TTRL~\citep{zhang2025no,he2026far}. In contrast, our design avoids interference across samples by isolating experience at the level of individual memories, enabling specialization and selective retrieval.
Finally, we note that this modular memory approach is supported by recent position work~\citep{dorovatas2026modular}, which outlines a framework for continual learning via separately accumulated memories that may later be merged or consolidated into the main model during dedicated offline phases. We leave this direction for future work.



\subsection{How to Leverage Latent Memories for Elevating Test-Time Performance}

Having described how to efficiently construct a memory that captures the experience gained from solving a given inquiry, we now turn to the question of how to reliably and effectively leverage previously accumulated memories at test time. We assume a pool of learned memories
\[
\mathcal{M} = \{ m^{(i)} \}_{i=1}^{N},
\]
constructed from previously encountered test samples $\{x^{(i)}\}$.
\\
\myparagraph{Memory Key Construction.}
For each memory $m^{(i)}$, we construct an associated key $k^{(i)}$ to enable retrieval. Owing to the latent nature of our memories, we consider two design choices for key representation. First, keys can be derived by pooling directly over the learnable token embeddings of the soft prompt. Alternatively, the soft prompt can be passed through the model’s layers, and the resulting activations can be pooled to produce layer-wise keys.
\\
\myparagraph{Memory Retrieval.}
Given a new input, we retrieve the most relevant stored memory by computing cosine similarity between a query vector derived from the current input and the keys associated with each stored memory. We consider two retrieval mechanisms.
\\
\textbf{Input retrieval} derives both queries and keys from the model’s input embedding space. Given an input $x$, we compute a query vector $q = \operatorname{mean}_t(\mathbf{E}(x)) \in \mathbb{R}^D
$, 
where $\mathbf{E}(x)$ denotes the input embedding matrix and the mean is taken over the sequence dimension. Retrieval then selects the memory with the highest cosine similarity:
\[
m^* = \operatorname{argmax}_{m^{(i)} \in \mathcal{M}} \operatorname{sim}(q, k^{(i)}),
\]
where $k^{(i)}$ is the key associated with memory $m^{(i)}$, constructed analogously at memory creation.
\\
\textbf{KV retrieval} operates in the model’s internal representation space. For each layer $l \in \{1, \dots, L\}$, we extract query activations from the QKV projections and pool across heads and sequence dimensions to obtain a layer-wise query $q^{(l)} \in \mathbb{R}^D$. Retrieval is performed independently at each layer using cosine similarity, and the resulting scores are averaged across layers to produce a final ranking:
\[
m^* = \operatorname{argmax}_{m^{(i)} \in \mathcal{M}} \frac{1}{L} \sum_{l=1}^{L} \operatorname{sim}\big(q^{(l)}, k^{(i,l)}\big),
\]
where $k^{(i,l)}$ denotes the layer-$l$ key associated with memory $m^{(i)}$. During memory creation, we extract and pool the key vectors from the KV matrices at each layer, yielding the corresponding layer-specific key $k^{(i,l)}$.
\\
\myparagraph{Response Routing.}
Because memories are learned using self-supervised signals, they can be noisy and may occasionally degrade performance. As we show in our offline analysis, although individual memories improve performance on average, they can sometimes harm specific inputs by turning a correct zero-shot response into an incorrect one. Ensuring reliability is therefore a key requirement of our online protocol, and our modular design naturally enables a lightweight mitigation strategy. Specifically, given a test sample $x$, we retrieve the nearest memory and generate both a memory-augmented response $a_{\text{mem}}$ and a zero-shot response $a_{\text{zs}}$. If they agree, we return the shared answer; otherwise, a verifier selects the higher-scoring response, mitigating memory-induced regressions. \\
Algorithm~\ref{alg:online_memory} outlines the main steps of our full  method, termed \ours  (Experiential Latent Memory).

\section{Experiments} 
This section validates the following questions:
(1) Can training with internal rewards on individual samples yield generalization performance?
(2) Is our memory parameterization effective?
(3) Can ELM achieve significant gains over training-free baselines?

\myparagraph{Experimental Setup.}
We  present the experimental setup across both offline and online settings.\\
\myparagraph{Benchmarks.}
We evaluate on three challenging mathematical reasoning benchmarks. \textbf{MATH500}~\citep{lightman2023let} is a 500-problem subset of the MATH dataset~\citep{hendrycks2measuring}, spanning competition-level problems across algebra, geometry, number theory, and related topics. \textbf{AMC23}~\citep{li2024numinamath} and \textbf{AIME24}~\citep{aime24} comprise problems from the 2023 AMC and 2024 AIME competitions respectively, representing challenging olympiad-level reasoning problems. 

\myparagraph{Models.}
We evaluate on two model families spanning complementary axes of the design space. \textbf{Llama-3.1-8B-Instruct}~\citep{grattafiori2024llama} is a general-purpose instruction-tuned model with no domain-specific mathematical training, while \textbf{Qwen2.5-Math-7B}~\citep{yang2024qwen2} is a math-specialized base model. This pairing allows us to assess our method across both domain-general vs.\ math-specialized pretraining, and instruction-tuned vs.\ base model settings, providing a broad and complementary evaluation spectrum.
\\
\myparagraph{ Implementation Details.} For individual sample training, we apply 10 gradient steps per example. We use 8 rollouts per step and a maximum sequence length of 2048, without up-scaling the number of samples for majority voting reward estimation (unlike~\citep{zuo2025ttrl})~\footnote{We note that this serves as a constrained instantiation of the original TTRL setup which uses 64 samples for extracting the label and 32 for training. However, follow-up work~\citep{he2026far} showed that 8 rollouts yields stable training.}. In terms of \textbf{evaluation metrics}, we report greedy \textit{pass@1}. Each soft prompt memory consists of 20 trainable tokens initialized from the embeddings of the input question. When ablating  LoRA based memories, we use rank $r=16$ applied to all attention and MLP projections. Refer to Appendix~\ref{appendix:exp_details} for full implementation details.

\begin{wrapfigure}{r}{0.49\textwidth}
    \vspace{-0.4cm}
    \centering
    \includegraphics[width=0.49\textwidth]{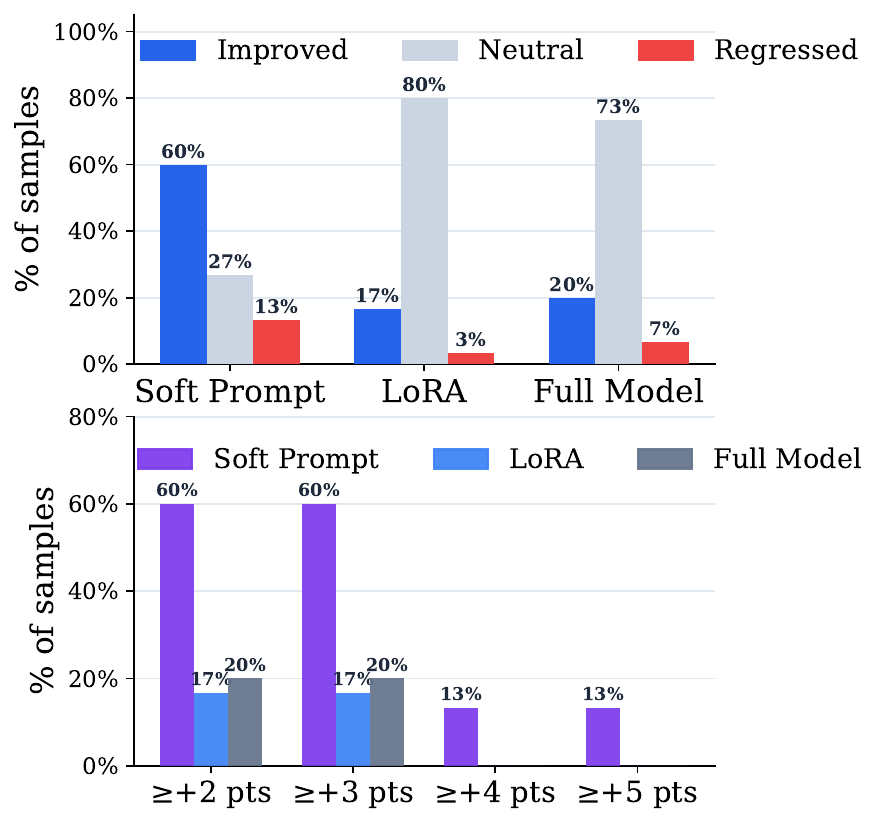}
    \caption{\small Memory design analysis on AIME24 with Llama. \textit{Top}: \% of training samples improving over zero-shot. \textit{Bottom}: \% exceeding score improvement thresholds.}
    \label{fig:offlinestats}
    \vspace{-0.3cm}
\end{wrapfigure}

\subsection{Offline Experiments}

\begin{table*}[t]
\centering
\small
\setlength{\tabcolsep}{6pt}

\begin{tabular}{l c c c c}
\toprule
& \#Train Params & MATH500 & AMC23 & AIME24 \\
\midrule

\multicolumn{5}{l}{\textbf{Llama-3.1-8B-Instr.}} \\

\rowcolor{gray!10} Zero-shot CoT & -- & 45 & 20 & 3.3\\
\addlinespace[2pt]

\rowcolor{gtgreen} GRPO & 8B & 47 & 27.5 & 6.7\\
\rowcolor{groupgreen} TTRL & 8B & 45.6 & 27.5 & 10\\

\addlinespace[3pt]

\rowcolor{groupblue} \ours$_\text{offline}$ & \textbf{0.08M} & 45 & \textbf{32.5} & \textbf{10}\\
\rowcolor{groupblue} ↳ oracle ZS vs Mem & -- &  55.4 & 35 & 10 \\
\rowcolor{groupblue} \ours-Full Model$_\text{offline}$ & 8B & 46 & 27.5 & 6.7\\
\rowcolor{groupblue} \ours-LoRA$_\text{offline}$ & 42M & 46.2 & 30 & 6.7\\

\midrule

\multicolumn{5}{l}{\textbf{Qwen2.5-Math-7B}} \\

\rowcolor{gray!10} Zero-shot CoT & -- & 67 & 50& 10\\
\addlinespace[2pt]

\rowcolor{gtgreen} Full GT & 7B & 68 & 62.5 & 20 \\
\rowcolor{groupgreen} Full Maj & 7B & 67.2 & 62.5 & 20\\

\addlinespace[3pt]

\rowcolor{groupblue} \ours$_\text{offline}$ & \textbf{0.07M} & \textbf{73.4} & \textbf{65} & \textbf{20}\\
\rowcolor{groupblue} ↳ oracle ZS vs Mem & -- & 77.6 & 67.5 & 22.5\\
\rowcolor{groupblue} \ours-Full Model$_\text{offline}$ & 7B &  67.8 & 60 & 13.33\\
\rowcolor{groupblue} \ours-LoRA$_\text{offline}$ & 40M & 68.6 & 62.5 & 10\\

\bottomrule
\end{tabular}

\caption{Accuracy (greedy pass@1) across benchmarks when trained with TTRL, , i.e., GRPO using a majority voting (Maj) reward. \colorbox{groupgreen!60}{Green} rows correspond to full-dataset training, while \colorbox{groupblue!60}{Blue} rows correspond to per-sample (modular) training. We also show results with ground truth (referred to as GRPO) rewards for the full model full-dataset training case.}
\label{tab:offline_tab}
\vspace{-0.5cm}
\end{table*}

To assess the generalization potential of training on individual samples with internal rewards, we introduce the following offline experimental protocol:
for each sample in the evaluation set, we train a latent memory and then evaluate it on the full set. AMC23 and AIME24 are small enough to allow  per-sample training across all examples; for these datasets, we run each experiment with three seeds to mitigate variance. MATH500 is  expensive to  evaluate exhaustively; instead, we sample multiple subsets using different seeds, train per-sample memories on each subset, and select the best-performing memories for evaluation on the full set.
This setup corresponds to an offline variant of our method without retrieval or verification, which we denote as \ours$\text{offline}$. 

As a comparison point, we also perform full-dataset offline training using either ground-truth rewards (with GRPO) or majority voting (TTRL) on the full model. Training is performed for 10, 30, and 80 epochs on MATH500, AMC23, and AIME24, respectively, following the setup of~\citep{zuo2025ttrl}. To ablate the choice of memory parameterization, we consider a LoRA-based parameterization~\citep{hu2022lora} (\ours-LoRA$\text{offline}$), as well as full-model per-sample training (\ours-Full Model$_\text{offline}$). For the per-sample variants, we report the average performance across seeds of the best-performing memory.

\begin{wraptable}{r}{0.32\textwidth} 
\vspace{-0.4cm}
\centering 
\begin{tabular}{l c | c} 
& Flip & Overlap \\ 
Model & 1$\rightarrow$0 & IoU \\ \hline 
Qwen & 18.5\% & 0.55 \\ 
LLaMA & 21.6\% & 0.37 \\ \hline
\end{tabular} \caption{\small Negative flip rate (1$\rightarrow$0) for top-5 memories and IoU  between top-2 memories on AMC23.} \label{tab:combined} 
\vspace{-0.2cm}
\end{wraptable}
Results are reported in Tab.~\ref{tab:offline_tab}. As observed, training on individual samples is highly effective across models and datasets, often yielding performance that exceeds full-dataset training, even with ground-truth rewards (GRPO).
Furthermore, among memory parameterizations, latent memories yield the most consistent generalization gains despite representing only $\sim0.001\%$ of model parameters. In contrast, full-model parameterization generalizes poorly, likely due to a mismatch between capacity and the limited training regime (single sample, 10 steps). 
Fig.~\ref{fig:offlinestats} and~\ref{fig:appendix_offline_stats} analyze these behaviors for AMC23 and AIME24: the top row shows the fraction of training samples that improve over zero-shot on the full evaluation set, while the bottom row reports the fraction exceeding given improvement thresholds. Overall, latent memories achieve the most consistent gains and are the only representation surpassing +3 points on the challenging AIME24 benchmark~\footnote{We omit LoRA and full-model results on MATH500 due to the substantially higher computational cost at this scale; comparisons are consistent with trends observed on AMC23 and AIME24.}\\
\myparagraph{Memory Specialization.} Tab.~\ref{tab:offline_tab} reports, for \ours$_\text{offline}$, a variant of oracle selection between the zero-shot response and the memory-aided response for each test sample during eval, showing that it further improves overall performance. This serves as evidence motivating our retrieval and response routing components: although individual memories improve performance on average, they can degrade specific samples by turning a correct zero-shot prediction into an incorrect one. \\
To investigate this phenomenon, we show in Tab.~\ref{tab:combined} that the percentage of negative flips\footnote{This rate is defined as $(\#\text{ negative flips}) / (\#\text{ negative flips} + \#\text{ positive flips})$} for the top-5 memories on AMC23 is around $20\%$ across models. In contrast, the Intersection over Union (IoU) between samples improved by the top-2 memories is moderate for both models. This indicates that each memory tends to improve a different subset of test samples, with only partial overlap.\\
This suggests that memories exhibit \textit{sample-dependent specialization}, rather than a single memory being uniformly optimal. Consequently, effective retrieval is necessary to match test queries to appropriate memories, and a robust routing mechanism is required to reliably select between zero-shot and memory-augmented predictions in the online setting.
\subsection{ Online Experiments}
\label{sec:online_experiments}
\begin{wraptable}{r}{0.46\textwidth}
\vspace{-0.5cm}
\centering
\small
\setlength{\tabcolsep}{3.pt}

\begin{tabular}{lccc}
\toprule
 & MATH500 & AMC23 & AIME24 \\
\midrule

\multicolumn{4}{l}{\textbf{Llama-3.1-8B-Instr.}} \\
\rowcolor{gray!10}
Zero-shot CoT & 45 & 20 & 3.3 \\

\addlinespace[2pt]
\rowcolor{groupgreen}
\multicolumn{4}{l}{\textit{Online}} \\
ICL & 46.3 & 21.7 & 3.3 \\
Reasoning Bank & 42.7 & 24.2 & 3.3\\
LAG & 44 & 19.2 & 4.4 \\
\ours$_{\text{input}}$ & \textbf{51.4} & \textbf{26} & 6.7\\
\ours$_{kv}$ & 50.7& 23.3 & \textbf{7.8}\\

\rowcolor{groupblue}
\multicolumn{4}{l}{\textit{Offline}} \\
GRPO & 47 & 27.5 & 6.7\\
TTRL & 45.6 & 27.5 & 10\\

\midrule

\multicolumn{4}{l}{\textbf{Qwen2.5-Math-7B}} \\
\rowcolor{gray!10}
Zero-shot CoT & 67 & 50 & 10 \\

\addlinespace[2pt]
\rowcolor{groupgreen}
\multicolumn{4}{l}{\textit{Online}} \\
ICL & -- & 38.3 & 12.2\\
Reasoning Bank & -- & 37.5 & 6.7 \\
LAG & -- & 40 & \textbf{16.7} \\
\ours$_{input}$ & 72.5 & 56 & 12\\
\ours$_{kv}$ & \textbf{72.6} & \textbf{62.5} & \underline{13} \\

\rowcolor{groupblue}
\multicolumn{4}{l}{\textit{Offline}} \\
GRPO & 68 & 62.5 & 20 \\
TTRL & 67.2 & 62.5 & 20 \\

\bottomrule
\end{tabular}

\caption{\textbf{Main Online Results.} Performance comparison (greedy pass@1) of \ours against training-free ICL baselines and full-offline training across datasets.}
\label{tab:online_main}
\vspace{-0.4cm}
\end{wraptable}
Now we turn our focus to the primary goal of this work: a strictly online streaming scenario where the model encounters inputs one at a time, in order, with no prior access to the evaluation data. This distinguishes our setting from two commonly conflated paradigms: a) \textit{test-time adaptation~\citep{wang2020tent}}, which trains  a model on a sample before the evaluation on the sample itself, and b) \textit{unsupervised RL} methods (e.g., TTRL~\citep{zuo2025ttrl}), which  require the full test set upfront to generate training signal before evaluating on the same set. 

\myparagraph{Baselines.} We compare against three experiential memory approaches: 
\textbf{(1) ICL} stores raw question--solution pairs $(q, \text{CoT})$ and retrieves the most similar example at test time to prompt the model in context. 
\textbf{(2) Reasoning Bank}~\cite{ouyang2025reasoningbank} solves each problem, performs a correctness-checking pass, and then a refinement pass to extract reusable reasoning strategies, which are stored in memory and retrieved to augment future queries. 
\textbf{(3) LAG}~\cite{chen2025log} uses latent representations extracted from the key-value (KV) activations of prior computations, retrieves them based on their textual representation, and prepends them to the query. 
For reference, we also report offline full-dataset training methods, namely GRPO and TTRL.~\footnote{To ensure a fair comparison, in Appendix~\ref{appendix:additional_comparisons}, we report results for a zero-shot baseline using two sampled responses with verifier-based selection, matching the additional compute introduced by our routing mechanism}. We use a Process Reward Model (PRM) as the verifier,  Qwen2.5-Math-PRM-7B~\citep{zhang2025lessons}.\\
As shown in Tab.~\ref{tab:online_main}, \ours is the strongest online method, consistently matching or surpassing full offline training. In contrast, training-free ICL baselines yield inconsistent gains, often improving one dataset while degrading another. Across all benchmarks, \ours delivers stable improvements over zero-shot performance and frequently closes the gap to offline approaches. On MATH500, \ours builds memories entirely online---without prior dataset access---yet still outperforms both GRPO and TTRL. This setting is particularly representative because the longer stream enables effective memory accumulation, while it is the setting where unsupervised RL (TTRL) start to degrade~\citep{he2026far}. Even on the shorter AMC23 and AIME24 benchmarks, \ours substantially improves base performance ($>+5$ points); notably, Qwen on AMC23 matches offline baselines while achieving a +12.5 point gain over the base model. In terms of retrieval mechanisms, both consistently outperform the baseline and are comparable with small differences across models and datasets. ~\footnote{Due to large dataset size, full results for Qwen-Math500 in the training-free settings were not available at the time of submission due to incomplete runs.}
\section{Ablations}
In this section we ablate several design choices of our proposed method, specifically the training objective, memory design and the potential of transfer of memories across datasets. \\
\myparagraph{Memory Design: Soft Prompts vs.\ LoRA.} Although LoRA is a widely used method for parameter-efficient adaptation, our results indicate that soft prompts are better suited for \ours. In the offline setting (Table~\ref{tab:offline_tab}), soft prompts consistently outperform LoRA in generalization. 
We further ablate LoRA in the online setting. For \textbf{LoRA} memories, keys are obtained by passing the input question through the model and extracting activations; these are computed once at memory creation and stored with the adapter. As shown in Appendix Table~\ref{tab:lora_onlne}, LoRA yields no improvement over the zero-shot baseline. 
Sec.~\ref{appendix:lora_retr} further shows that top-retrieved LoRA memories tend to be those that also perform poorly offline, indicating a retrieval misalignment. Soft prompts naturally avoid this issue, as they are trained in the embedding space and thus remain aligned with the retrieval mechanism. 
\begin{wrapfigure}{r}{0.33\textwidth}
    \vspace{-0.5cm}
    \centering
    \includegraphics[width=0.33\textwidth]{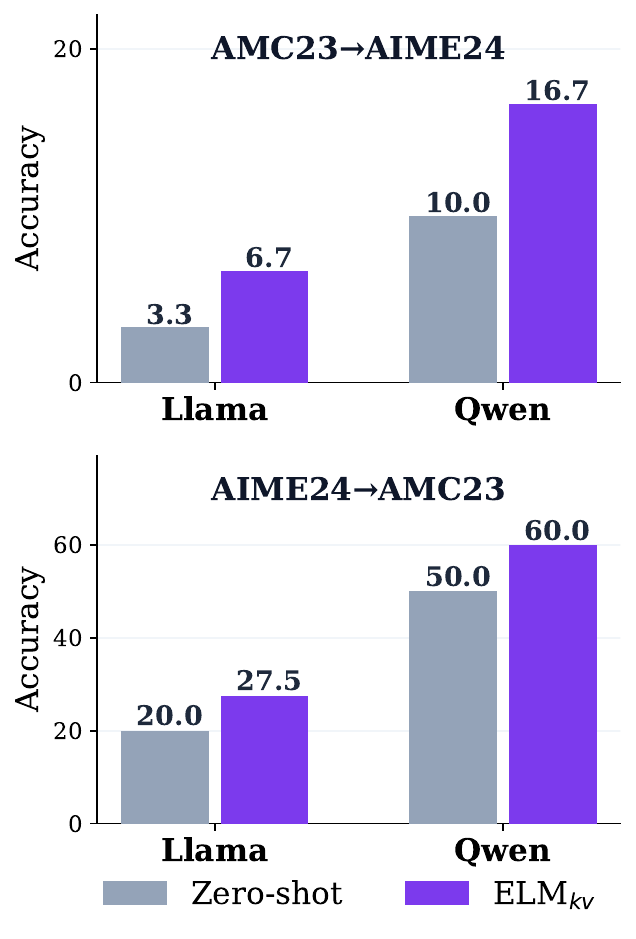}
    \caption{\small Cross-dataset transfer of \ours memories.}
    \label{fig:transfer}
    \vspace{-0.6cm}
\end{wrapfigure}
Beyond performance, soft prompts are also more compute- and memory-efficient, train faster, and incur lower inference overhead (Appendix~\ref{appendix:lora_vs_sp}).\footnote{We also discuss Prefix Tuning as an alternative in Appendix~\ref{appendix:prefix}.} \\
\myparagraph{Does \ours transfer across datasets?} Since our main evaluation runs ELM online on test samples from a given dataset, we examine whether the resulting memories transfer across datasets. Table~\ref{fig:transfer} shows transfer between AIME24 and AMC23, using pre-constructed memories from the other set,  compared to the zero-shot baseline. Notably, we observe consistent gains on the target dataset, indicating that the learned memories generalize beyond a single test set.
\\
\myparagraph{Objective.} We use GRPO with majority voting, converting self-generated outputs into rewards. Unlike SFT on greedy responses, which can reinforce errors in the absence of ground truth, GRPO exploits additional test-time compute to produce stronger training signals, leading to more robust generalization~(Tab.~\ref{tab:ablations}).\\
\myparagraph{Effect of Training Steps.} In the main experiments, we train memories for 10 steps, as a good trade-off between convergence and training time. In Tab.~\ref{tab:ablations}, we report results with only 2 steps, showing smaller but consistent gains over the zero-shot baseline.

\section{Discussion and Limitations}
Given the strong gains achieved by large reasoning models with increased test-time compute, we study how to efficiently distill such gains into reusable experience that improves future performance at lower cost. We introduce Experiential Latent Memory (ELM), a lightweight modular memory learned per sample using GRPO with internally derived rewards. We show that individual memories can generalize to unseen data and approach the performance of full offline training. However, memory quality varies, and some memories are noisy and can degrade otherwise correct responses.
To address this, we leverage the modularity of ELM and propose an online retrieval protocol that augments test-time inference with relevant memories. To ensure reliability, we employ a verifier that selects between zero-shot and memory-augmented responses. This design yields strong performance gains with minimal overhead and enables efficient reuse of experience. Overall, this work represents a first step toward models capable of continuous self-improvement. \\
Several limitations remain. While we demonstrate the utility of individually trained memories, we do not address how to identify samples that lead to high-quality memories. Preliminary experiments show no strong correlation between simple training-free statistics and memory utility. Although retrieval filters many irrelevant memories, some retrieved memories can still be harmful, likely due to incorrect self-supervision; we therefore rely on a verifier and leave the design of more efficient decision strategies to future work. Due to computational constraints, we limit our analysis of what individual memories encode and of alternative objectives beyond majority voting. Finally, important open questions remain concerning memory compositionality, as well as how memories should be updated, merged, evicted, or distilled into the model at later stages. We hope this work provides a foundation for reliable and efficient continual self-improvement.
\newpage

{
\small
\bibliographystyle{plain}
\bibliography{neurips_2026}
}
\newpage
\appendix

\section{Additional Experimental Details}
\label{appendix:exp_details}

All models are trained using the TRL GRPO Trainer (Hugging-Face) with the DR-GRPO~\citep{liu2025understanding} loss variant. As mentioned, we use a group size of $G{=}8$ completions per prompt, a maximum completion length of 2048 tokens, and a KL penalty of $\beta{=}0.01$. We use the AdamW fused optimizer with a constant learning rate of $10^{-4}$ for soft prompt and LoRA memories, $10^{-3}$ for Prefix Tuning and $10^{-6}$ for full model training, with a global batch size of 64 and no gradient accumulation. For Qwen models we use a sampling temperature of $0.8$ and $0.7$ for Llama. All training is performed on a single H200 GPU, with evaluations conducted on an A100 GPU using the \texttt{lm-eval-harness} library. The offline experiments on AMC23 and AIME24 require around 12 hours to complete, while for online, extracting memories (training) requires around 5 hours for AMC23 and AIME24, and 1 day for MATH500, and then evaluation needs 2h for the small ones and around 10 hours for MATH500. For the \textit{offline} experiment on MATH500 subsets we use the seeds $[17,42,123,456,789,1234,5678,9999]$, with a subset of 80 which we split into 40 for training and 40 for evaluation. For the \textit{online} experiments we use 3 seeds, namely $[17, 42, 78]$ and we report the mean over them.\\
\myparagraph{Online Experiments.}We always retrieve the top-1 memory for each input. The verifier model we use is a Process Reward Model (PRM), specifically Qwen2.5-Math-PRM-7B~\citep{zhang2025lessons}. Finally, for each online experiment we run 3 seeds and we report average performance for each method.
\subsection{Prompts}
We use a simple zero-shot chain-of-thought prompt across all experiments:
\begin{verbatim}
Please reason step by step, and put your final answer within \boxed{answer}.
{question}
\end{verbatim}
For Llama-3.1-8B-Instruct, this prompt is wrapped using the model's default chat template.

\section{Prefix Tuning Parametrization}
\label{appendix:prefix}

For completeness, we also compare against Prefix Tuning~\cite{li2021prefix} as an additional memory parameterization. Prefix Tuning prepends trainable continuous vectors directly to the keys and values of each transformer attention layer, offering a deeper form of conditioning than soft prompts. We initialize them from the key-value activations of the question (similar idea with the soft prompts). As shown in Table~\ref{tab:prefix}, Prefix Tuning underperforms  overall soft prompts, while being more parameter-heavy, further supporting our choice of soft prompt memories for \ours.

\section{Additional Offline Statistics}

\begin{figure}
    \centering
    \includegraphics[width=1\textwidth]{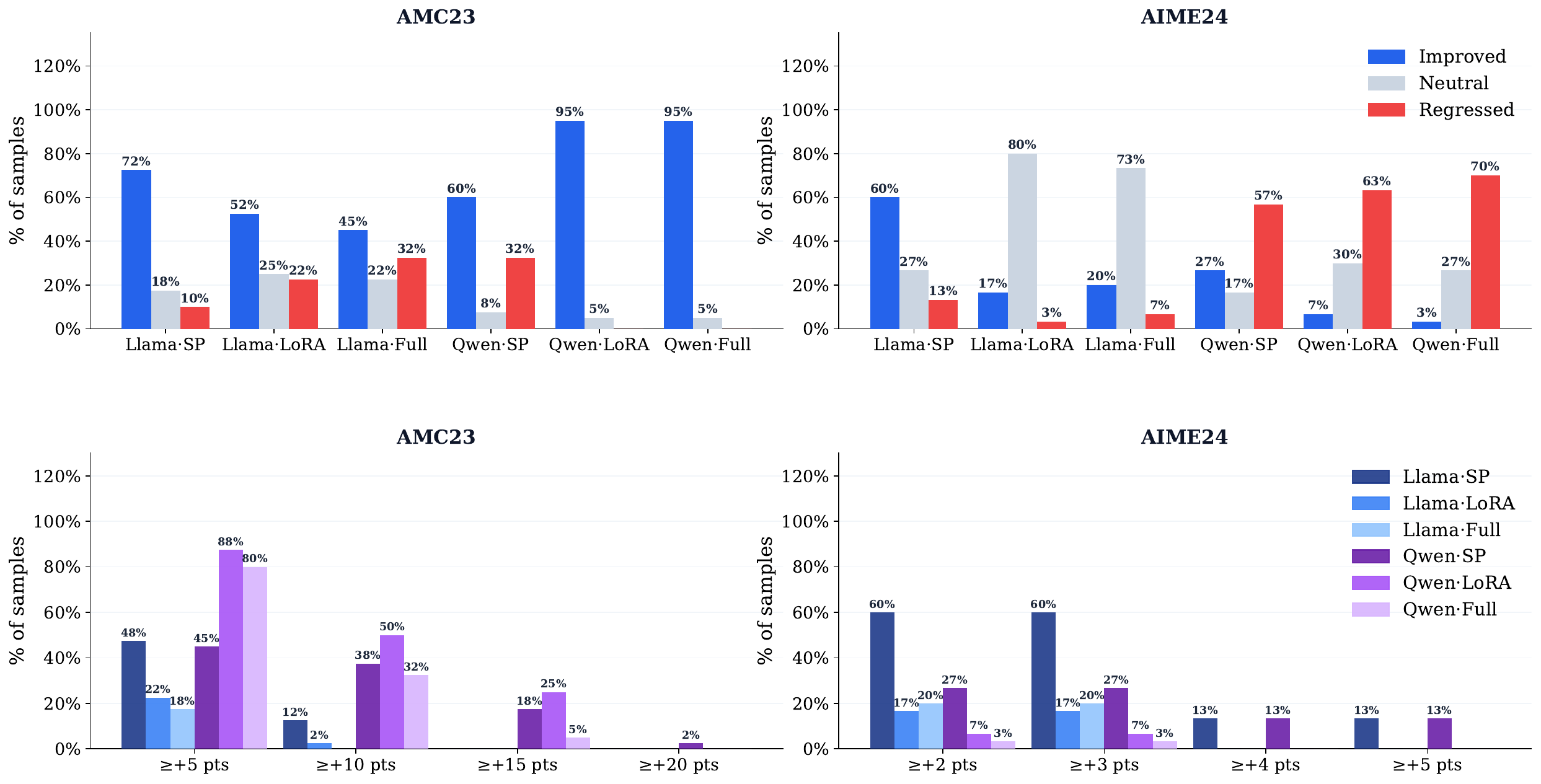}
    \caption{Statistics of \ours on ACM23 and AIME24 across memory designs.}
    \label{fig:appendix_offline_stats}
\end{figure}

We extend here Fig.~\ref{fig:offlinestats}
 of the main paper to include both models and AMC23. Results are in Fig.~\ref{fig:appendix_offline_stats}.
 
 \begin{table*}[t]
\centering
\small
\setlength{\tabcolsep}{6pt}

\begin{tabular}{l c c c}
\toprule
& \#Trainable Params & AMC23 & AIME24 \\
\midrule

\multicolumn{4}{l}{\textbf{Llama-3.1-8B-Instr.}} \\
\addlinespace[1pt]
\ours-SoftPrompt$_\text{offline}$ & {0.08M} & 32.5 & 10\\
\ours-LoRA$_\text{offline}$ & 42M & 30& 6.7 \\
\ours-PrefixTuning$_\text{offline}$ & 6.5B & 30 & 10\\

\midrule

\multicolumn{4}{l}{\textbf{Qwen2.5-Math-7B}} \\

\addlinespace[1pt]
\ours-SoftPrompt$_\text{offline}$ & {0.08M} & 65 & 20\\
\ours-LoRA$_\text{offline}$ & 42M & 62.5& 10 \\
\ours-PrefixTuning$_\text{offline}$ & 4M & 55 & 13.3 \\

\bottomrule
\end{tabular}

\caption{\textbf{Memory Design.} Average accuracy (greedy pass@1) across 3 seeds on AMC23 and AIME24.}
\label{tab:prefix}
\end{table*}

\section{Additional Comparison of Soft Prompts against LoRA}
\label{appendix:lora_vs_sp}

\subsection{Performance of LoRA in the online setting}

\begin{table*}[t]
\centering
\small
\setlength{\tabcolsep}{6pt}

\begin{tabular}{l c c}
\toprule
 & AMC23 & AIME24 \\
\midrule

\multicolumn{3}{l}{\textbf{Llama}} \\
\addlinespace[1pt]
↳ ZS & 20 & 3.3\\
\ours-LoRA  & 20 & 3.3\\

\midrule
\multicolumn{3}{l}{\textbf{Qwen}} \\

\addlinespace[1pt]
↳ ZS & 50 & 10\\
\ours-LoRA  & 50 & 10\\

\bottomrule
\end{tabular}

\caption{Average performance of \ours with LoRA memory parametrization (greedy pass@1) across 3 seeds on AMC23 and AIME24. As we discuss (Misaligned Retrieval), the top-retrieved memories in the online experiment are the ones that do not improve over base in the offline experiment, thus we see no improvement over baseline here.}
\label{tab:lora_onlne}
\end{table*}

\label{appendix:lora_retr}
\paragraph{Misaligned Retrieval.} To further understand the online failure of LoRA memories, we analyze retrieval statistics on AMC23 with Qwen. The three most frequently retrieved memories account for roughly 72\% of all retrievals: M13 (36\%), which is neutral in the offline setting with no improvement over zero-shot; M10 (23\%), which improves only a single sample; and M24 (13\%), which regresses in the offline setting. This confirms that the retrieval mechanism systematically favors memories that fail to generalize, highlighting a fundamental misalignment between retrieval and utility that explains the lack of online improvement observed for LoRA.

\subsection{Compute and Memory Overhead}

In the following table we show the number of trainable parameters, as well as the seconds per GRPO step of LoRA and Soft Prompts. Soft Prompt memories are significantly lighter. 

\begin{table}[h]
\centering
\small
\setlength{\tabcolsep}{4pt}
\begin{tabular}{l c c}
\toprule
 & \# Train Params & Train Latency (s/step)  \\
\midrule
\multicolumn{3}{l}{\textbf{Llama}} \\
\addlinespace[1pt]
↳ \ours & 0.08M & 140 \\
↳ \ours-LoRA & 42M & 70 \\
\midrule
\multicolumn{3}{l}{\textbf{Qwen}} \\
\addlinespace[1pt]
↳ \ours & 0.07M & 48 \\
↳ \ours-LoRA & 40M & 91 \\
\bottomrule
\end{tabular}
\vspace{0.1cm}
\caption{Efficiency comparison of Soft Prompts vs.\ LoRA memory design.}
\label{tab:efficiency}
\end{table}

\section{Additional Comparisons}
\label{appendix:additional_comparisons}

\subsection{Objective and Training Steps}

Due to space limitations in the main paper we present here the ablation table of traininig obective and number of training steps. 
\begin{table}[t]
\vspace{-0.2cm}
\centering
\small
\setlength{\tabcolsep}{3pt}  
\begin{tabular}{l|cc}
 & AMC23 & AIME24 \\
 \hline
 Zero-shot          & 50   & 10   \\
\hline
\ours         & 65   & 20   \\
\ours-2 S & 62.5 & 16.7  \\
\hline
\ours-SFT     & 62.5 & 13.3  \\
\end{tabular}
\caption{The effect of objective and only 2 training steps on \ours (with Qwen).}
\label{tab:ablations}
\vspace{-0.4cm}
\end{table}

\subsection{Zero-shot Baseline with \textit{verifier@2}}

To ensure fair comparison with our method, as mentioned in the main paper, here we present the results of the base model when sampling 2 responses and using the verifier to select between them, thus matching the extra computational overhead of our method. As we see in Tab.~\ref{tab:verif_at_2} this baseline elevates the performance of greedy zero-shot (as more test-time compute is spent) but on average lags behind \ours. 

\begin{table*}[t]
\centering
\small
\setlength{\tabcolsep}{6pt}

\begin{tabular}{l c c}
\toprule
 & AMC23 & AIME24 \\
\midrule

\multicolumn{3}{l}{\textbf{Llama}} \\
\addlinespace[1pt]
↳ ZS & 20 & 3.3 \\
\textit{verifier@2}  & 25 & 4.2\\

\midrule
\multicolumn{3}{l}{\textbf{Qwen}} \\

\addlinespace[1pt]
↳ ZS & 55 & 10\\
\textit{verifier@2}   & 57& 16\\

\bottomrule
\end{tabular}

\caption{Performance of base model zero-shot greedy vs \textit{verifier@2}.}
\label{tab:verif_at_2}
\end{table*}

\label{sec:offline_experiments}

\subsection{Creating training-free memories with additional test-time compute}

In the main paper, we evaluate training-free ICL methods that store either raw reasoning traces or latent representations derived from them as memories. These memories are constructed from greedy decoding traces. To ensure a fair comparison, and to further highlight the importance of training memories to achieve the appropriate level of abstraction for generalization, we additionally consider a stronger test-time baseline with the LAG method, where, after solving each input, we allocate extra inference-time compute via majority voting to construct the memory.

As shown in Table~\ref{tab:tts_mems}, despite the additional compute, the resulting memory representations still fail to generalize effectively. This further motivates our lightweight training pipeline, which enables experiences to be internalized into compact and reusable memories.

\begin{table*}[t]
\centering
\small
\setlength{\tabcolsep}{6pt}

\begin{tabular}{l c c}
\toprule
 & AMC23 & AIME24 \\
\midrule

\multicolumn{3}{l}{\textbf{Llama}} \\
\addlinespace[1pt]
ZS & 20 & 3.3\\
LAG & 19.2 & 4.4\\
LAG-TTS  & 14.2 & 3.3\\

\midrule
\multicolumn{3}{l}{\textbf{Qwen}} \\

\addlinespace[1pt]
ZS & 50 & 10\\
LAG & 40& 16.7 \\
LAG-TTS  & 36.7& 17.8\\

\bottomrule
\end{tabular}

\caption{LAG with additional test-time compute (majority voting) spent to construct memories.}
\label{tab:tts_mems}
\end{table*}


\end{document}